\documentclass[sigconf]{acmart}
\usepackage[ruled]{algorithm2e}
\usepackage{balance}
\AtBeginDocument{%
  \providecommand\BibTeX{{%
    \normalfont B\kern-0.5em{\scshape i\kern-0.25em b}\kern-0.8em\TeX}}}

\copyrightyear{2021}
\acmYear{2021}
\setcopyright{acmcopyright}
\acmConference[ICMR '21] {Proceedings of the 2021 International Conference on Multimedia Retrieval}{August 21--24, 2021}{Taipei, Taiwan.}
\acmBooktitle{Proceedings of the 2021 International Conference on Multimedia Retrieval (ICMR '21), August 21--24, 2021, Taipei, Taiwan}
\acmPrice{15.00}
\acmISBN{978-1-4503-8463-6/21/08}
\acmDOI{10.1145/3460426.3463588}

\begin{document}
\fancyhead{}

\title{NMS-Loss: Learning with Non-Maximum Suppression for Crowded Pedestrian Detection}

\author{Zekun Luo}
\affiliation{%
  \institution{Youtu Lab, Tencent}
  \city{Shanghai}
  \country{China}
}

\author{Zheng Fang}
\affiliation{%
  \institution{Beihang University}
  \city{Beijing}
  \country{China}
}

\author{Sixiao Zheng}
\affiliation{%
  \institution{Fudan University}
  \city{Shanghai}
  \country{China}
}

\author{Yabiao Wang}
\affiliation{%
  \institution{Youtu Lab, Tencent}
  \city{Shanghai}
  \country{China}
}

\author{Yanwei Fu}
\authornote{Yanwei Fu is with the School of Data Science and MOE Frontiers Center for Brain Science, Fudan University, Shanghai 200433, China (e-mail: yanweifu@fudan.edu.cn)}
\affiliation{%
  \institution{Fudan University}
  \city{Shanghai}
  \country{China}
}

\begin{abstract}

Non-Maximum Suppression (NMS) is essential for object detection and affects the evaluation results by incorporating False Positives (FP) and False Negatives (FN), especially in crowd occlusion scenes. In this paper, we raise the problem of weak connection between the training targets and the evaluation metrics caused by NMS and propose a novel NMS-Loss making the NMS procedure can be trained end-to-end without any additional network parameters. Our NMS-Loss punishes two cases when FP is not suppressed and FN is wrongly eliminated by NMS. Specifically, we propose a pull loss to pull predictions with the same target close to each other, and a push loss to push predictions with different targets away from each other. Experimental results show that with the help of NMS-Loss, our detector, namely NMS-Ped, achieves impressive results with Miss Rate of $5.92\%$ on Caltech dataset and $10.08\%$ on CityPersons dataset, which are both better than state-of-the-art competitors.

\end{abstract}




\maketitle

\section{Introduction}

Pedestrian detection ~\cite{dollar2009pedestrian} is an essential computer vision task that has numerous applications such as automatic driving, video surveillance and person re-identification. With the help of deep convolution neural networks (CNNs) ~\cite{zhang2015filtered, he2016deep, simonyan2014very}, the performance of pedestrian detection has been significantly improved. However, the False Negatives (FN) in crowd occlusion scenes and False Positives (FP) generated for the same person are still the fundamental challenges.

Existing methods for pedestrian detection can mainly be divided into two categories: hand-crafted feature based ~\cite{felzenszwalb2009object,felzenszwalb2010cascade,xu2014domain,yan2014fastest, dollar2009integral,dollar2014fast,nam2014local,zhang2015filtered} and deep learning based ~\cite{zhang2016faster,cai2016unified,brazil2017illuminating,du2017fused,ren2017accurate,mao2017can,wang2018repulsion,liu2018learning}. The first one applies the sliding-window way to get different scales of patches, then uses human-designed feature extractor such as Haar ~\cite{viola2001rapid} and HoG ~\cite{dalal2005histograms} to obtain feature representation, last utilizes SVM ~\cite{cortes1995support} classifier to filter background. These hand-crafted feature representations could not handle complex scenes. The second one uses deep convolutional neural networks (CNNs) to obtain high-level semantic feature representation, which has a discriminative ability to deal with complex scenes for pedestrian detection. To alleviate 
FN issue in high occlusion scenes, different variants of Non-Maximum Suppression (NMS) ~\cite{liu2019adaptive,huang2020nms, bodla2017soft} are proposed to change NMS threshold during inference adaptively. To reduce FP, many works ~\cite{chi2019relational,chi2020pedhunter} jointly predict pedestrian boxes and parts information such as head due to that it is less occluded. However, the objective between training and inference is inconsistent, which may result in sub-optimal performance for pedestrian detection. 

NMS is an essential procedure for object detection tasks. Modern pedestrian detectors rely on NMS to remove duplicate detections for both one-stage and two-stage approaches. The nearby detections around one object will be removed once its interaction over union (IoU) with the object is larger than the pre-defined threshold. During the training process, there is no such process, thus resulting in inconsistency between optimized detection training results and final inference results. To handle the inconsistency problem, NMS process should be incorporated into the training process. To this end, we propose a novel NMS-Loss. There are two components, pull and push losses, in our NMS-Loss. Pull loss aims to raise the precision by pulling FP close to the max score prediction, and push loss focuses on improving recall by pushing predictions away from each other. With the help of NMS-Loss, false predictions on the evaluation metric can be directly reflected on loss functions, and thus be directly optimized.

The main contribution of this work lies in the following aspects.
\begin{itemize}
    \item We firstly raise the problem of weak connection between training targets and evaluation metrics in pedestrian detection and propose a novel NMS-Loss making the NMS procedure can be trained end-to-end, which does not introduce any parameters nor runtime cost.
    \item We propose finely designed pull and push losses helping the network to boost performance on precision and recall, respectively, which considering both prediction coordinates and confidence.
    \item With the help of NMS-Loss, in pedestrian detection, our proposed NMS-Ped outperforms SOTA methods on the widely used Caltech and CityPersons datasets.
\end{itemize}

\section{NMS-Loss}
\subsection{Overview of NMS-Loss}

The traditional NMS procedure is shown in Alg.~\ref{nms_loss_alg} without considering the red texts. Starting with a set of detection boxes $\mathcal{B}$ with corresponding scores $\mathcal{S}$, NMS firstly moves the proposal $b_{m}$ with the maximum score from the set $\mathcal{B}$ to the set of final kept detections $\mathcal{K}$. It then removes any box in $\mathcal{B}$ and its score in $\mathcal{S}$ that has an overlap with the $b_{m}$ higher than a manually set threshold $N_{t}$. This process is repeated for the remaining $\mathcal{B}$ set.

However, no existing approaches take the NMS into the training process to adjust the detection boxes, making the learning targets inconsistent with the evaluation metric, which means FP not suppressed by NMS and FN eliminated by NMS can harm the precision and recall, respectively. To avoid inconsistency, we propose the NMS-Loss taking the NMS procedure into the training process, which adaptively selects the false predictions caused by NMS and uses two well-designed \emph{pull} and \emph{push losses} to minimize the FP and FN, respectively. Specifically, our NMS-Loss is defined as:
\begin{equation}
L_{nms}\ = \lambda_{pull}L_{pull} + \lambda_{push}L_{push},
\label{NMS_Loss}
\end{equation}

where $L_{pull}$ is the pull loss to punish the FP not suppressed by NMS and $L_{push}$ is the push loss to punish the FN wrongly eliminated by NMS. Coefficients $\lambda_{pull}$ and $\lambda_{push}$ are the weights for balancing losses. Details of our NMS-Loss are present in Algorithm \ref{nms_loss_alg} emphasized with red color. Different from the traditional NMS, we use a set $\mathcal{G}$ containing corresponding ground truth indexes of detection boxes, which is used to identify FP and FN. In the NMS-Loss calculating procedure, $\mathcal{M}$ is an auxiliary dictionary with the ground truth index as key and corresponding max score detection as value, which is used to record the max score prediction of each ground truth. Our NMS-Loss is naturally merged into the NMS procedure without incorporating any additional training parameters. The runtime cost of NMS-Loss is zero for testing.

\begin{algorithm}[t!]
\small
\SetAlgoLined
\KwIn{ \\
       \quad$\mathcal{B} = [b_{1},\ldots,b_{N}]$, $\mathcal{S} = [s_{1},\ldots,s_{N}]$, $N_{t}$, {\color{red}$\mathcal{G} = [g_{1},\ldots,g_{N}]$}\\
       \quad$\mathcal{B}$ is the list of initial detection boxes \\
       \quad$\mathcal{S}$ contains corresponding detection scores \\
       \quad$N_{t}$ is the NMS threshold \\
       \quad{\color{red}$\mathcal{G}$ contains corresponding ground truth indexes} \\
    }
\SetKwInput{AuxVar}{Auxiliary Variable}
    \AuxVar{
        \\ \quad$\mathcal{K} \gets [\ ]$, {\color{red}$\mathcal{M} \gets dictionary()$,} \\
        \quad$\mathcal{K}$ is the list to keep final detections after NMS \\
        \quad{\color{red}$\mathcal{M}$ is a dictionary using the ground truth index as key \\
         \quad and corresponding max score detection as value}
    }
\SetKwBlock{Begin}{begin}{end}
    \Begin{
        \While{$\mathcal{B} \neq empty$}{
            $m \gets argmax\ \mathcal{S}$ ; \\
            {\color{red}
            \eIf{$g_{m}\ not\ in\ \mathcal{M}.keys()$}{
                $\mathcal{M}[g_{m}] \gets b_{m}$; \\
            }{
                $b_{max} \gets \mathcal{M}[g_{m}]$; \\
                 pull\_loss$(b_{max}, b_{m})$;\quad Eq.~(\ref{Pull_Loss}) \\
            }
            }
            $\mathcal{K} \gets \mathcal{K} \cup b_{m}$ ; $\mathcal{B} \gets \mathcal{B} - b_{m};$ \\
            $\mathcal{S} \gets \mathcal{S} - s_{m}$ ; {\color{red}$\mathcal{G} \gets \mathcal{G} - g_{m}$;} \\
            \For{$b_{i}\ in\ \mathcal{B}$}{
                \If{$IoU(b_{m}, b_{i})\ge N_{t}$}{
                    {\color{red}
                    \If{$g_{m} \neq g_{i}$}{
                        push\_loss$(b_{m}, b_{i})$;\quad Eq.~(\ref{Push_Loss}) \\
                    }
                    }
                    $\mathcal{B} \gets \mathcal{B} - b_{i}$; $\mathcal{S} \gets \mathcal{S} - s_{i}$; 
                    {\color{red}$\mathcal{G} \gets \mathcal{G} - g_{i}$;} \\
                }
            }
        }
        \Return{$\mathcal{K}$}
    }
 \caption{\textbf{NMS-Loss Calculating Procedure}}
 \label{nms_loss_alg}

\end{algorithm}

\begin{figure}[t!]
\centering
\includegraphics[clip=true,width=0.8\linewidth]{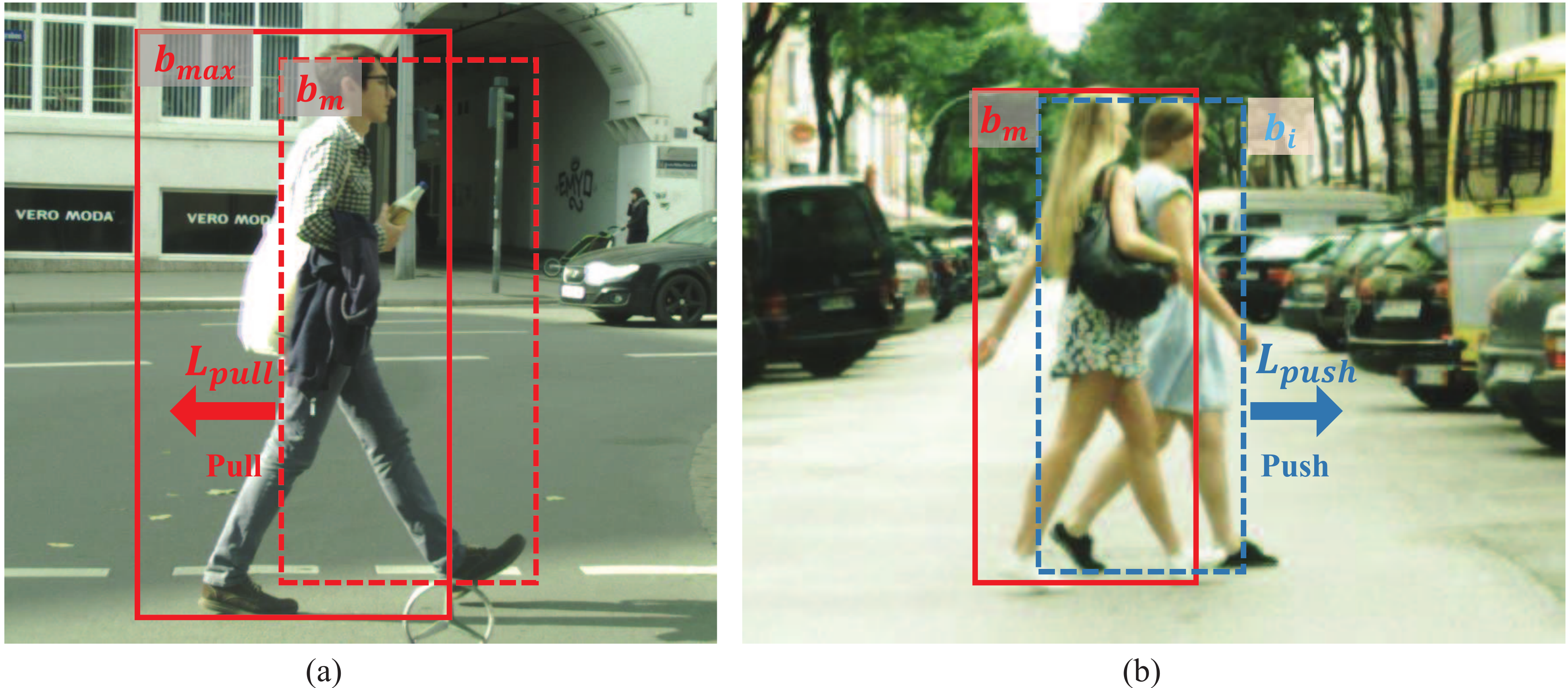}
\vspace{-3mm}
\caption{\textbf{Illustration of our NMS-Loss}. All boxes $b_{max}$, $b_{m}$ and $b_{i}$ are predictions as described in Alg.~\ref{nms_loss_alg}, where boxes with the same color have the same target and boxes with the solid line get a higher score than boxes with the dotted line. In (a), $b_{m}$ is a FP not suppressed by $b_{max}$. Our $L_{pull}$ pulls $b_{m}$ towards $b_{max}$. In (b), $b_{i}$ is a FN wrongly eliminated by $b_{m}$. Our $L_{push}$ pushes $b_{i}$ away from $b_{m}$.}
\vspace{-3mm}
\label{visible_nms_loss}
\end{figure}

\subsection{Pull Loss Definition}
With the objective to reduce FP, we need to find out wrongly kept predictions. To this end, in every iteration, we check whether the current max score prediction $b_{m}$ is the max score prediction for its corresponding $g_{m}$ ground truth. If not, it means $b_{m}$ is an FP not suppressed by NMS, pull loss should be performed between $b_{m}$ and the max score prediction $b_{max}$ of the $g_{m}$ ground truth (see Fig.~\ref{visible_nms_loss}). Formally, our pull loss is calculated as:
\begin{equation}
L_{pull}\ =\ -ln(1-N_{t}+IoU(b_{max},b_{m}))s_{m},
\label{Pull_Loss}
\end{equation}
where $N_{t}$ is the predefined NMS threshold and $s_{m}$ is the prediction score corresponding to $b_{m}$. We note two properties of the pull  loss: (1) When the IoU between $b_{max}$ and $b_{m}$ is small, pull loss tends to increase, forcing the network to learn to pull $b_{m}$ toward $b_{max}$. The NMS threshold $N_{t}$ is used to prevent the gradient of outliers influence too much on model learning. Besides, for the NMS procedure, we just need to make the IoU between FP and TP higher than $N_{t}$. Using $N_{t}$ in pull loss to reduce the gradient of outliers can make the network easy to learn. (2) The prediction score of FP can also have a strong effect on pull loss. FP with a higher score has a greater impact on evaluation results and intuitively needs to be paid more attention. Besides, it makes the network learn to fix FP not only just conditioning the box coordinates but also considering lower the prediction scores.

\subsection{Push Loss Definition}
In NMS, the current max score prediction $b_{m}$ eliminates boxes which get an IoU higher than $N_{t}$ with $b_{m}$. If the eliminated box $b_{i}$ corresponds to different ground truth index with $b_{m}$, $b_{i}$ will be a FN and reduce recall (see Fig.~\ref{visible_nms_loss}). To avoid $b_{i}$ from being wrongly eliminated, we propose a push loss to penalize FN:
\begin{equation}
L_{push}\ =\ -ln(1-IoU(b_{i},b_{m}))s_{i},
\label{Push_Loss}
\end{equation}
where $s_{i}$ is the prediction score corresponding to $b_{i}$. Different from pull loss, as $IoU(b_{i},b_{m}) \to 1$, the push loss goes higher and the model learns to push $b_{i}$ away from $b_{m}$. To avoid the model tending to reduce the push loss by lowering the score of FN, we use the $s_{i}$ only for reweighting losses without back propagating gradient.

For crowded scenes, especially in the CityPersons dataset, the ground truths of bounding boxes are overlapped with each other.
It is unreasonable to push their predictions away from each other with an IoU equals to zero. To handle this problem, we only calculate $L_{push}$ on prediction whose IoU is higher than the IoU of its corresponding ground truth boxes.

Our pull and push loss are performed on predictions. When the pull/push loss is activated, the network tries to pull/push both predictions close to/away from each other, respectively.
Since high score predictions generally get a more accurate location, it is unreasonable to move an accurate prediction based on an inaccurate one. To handle this, we stop the gradient backward propagation of high score predictions, leading the network to focus on false predictions.

\section{Experiments}
\subsection{Experimental Setup}

\noindent
\textbf{Datasets and Evaluation metrics.} We evaluate our method on two challenging pedestrian datasets: Caltech~\cite{dollar2009pedestrian,dollar2011pedestrian} and CityPersons~\cite{zhang2017citypersons}. We report performance using standard average-log MR between [$10^{-2},10^{0}$] of False Positive per Image (FPPI). A minimum IoU threshold of $0.5$ is required for detected box to match with a ground truth box. By default, we report the results on Reasonable subsets is a widely used setup where the pedestrian is at least $65\%$ visible and $50$ pixels tall.

\noindent
\textbf{Experimental Settings.}
As shown in RPN+BF~\cite{zhang2016faster}, small instances are hard to be detected in the low-resolution feature maps provided by RoI-Pooling, which is more severe in pedestrian detection. Therefore, we used Faster R-CNN~\cite{ren2015faster} as our baseline, but made two adjustments: (1) Inspired by~\cite{zhang2016faster}, we use a separate network to construct the RCNN and put the cropped original image to RCNN for further refinement. This improves the ability of the network to detect small instances, but it is not suitable for instances with large scale changes. (2) There is an additional weak semantic segmentation loss~\cite{brazil2017illuminating} to boost performance. Note that the baseline has the same settings as our NMS-Ped except that there is no NMS-Loss in baseline.

PyTorch~\cite{paszke2017automatic} is used to train the NMS-ped for both datasets. We use $8$ NVIDIA GPUs with a mini-batch comprises $1$ image per GPU. SGD with momentum of $0.9$ and weight decay of $1 \times 10^{-4}$ is adopted for training. Both datasets are trained only using the images with foreground. Random cropping and flipping are used for data augmentation. Detailed settings on Caltech and CityPersons are described as follows:

\textbf{Caltech:}
The learning rate for Caltech is $5$$\times$$10^{-3}$ and is dropped by a factor of $10$ after $9,600 $ iterations and $13,200$ iterations. The images are resized to $1280 \times 960$ in our experiments. The weights for pull loss and push loss are both $0.1$ getting from experiments.

\textbf{CityPersons:}
The learning rate for CityPersons is $1$$\times$$10^{-2}$ and dropped by a factor of $10$ after $24,000$ iterations and $33,000$ iterations. We use the original image resolution of $2048$$\times$$1024$ in our experiments. The weights for pull and push loss are $0.1$ and $0.001$ respectively for the reason that CityPersons contains much more crowded scenes than Caltech and lots of instances are heavily overlapped with others. Giving a relatively lower weight for push loss will reduce the gradient of pushing and make multi-tasks work well.

\begin{figure}[t]
\centering
\includegraphics[clip=true,width=0.8\linewidth]{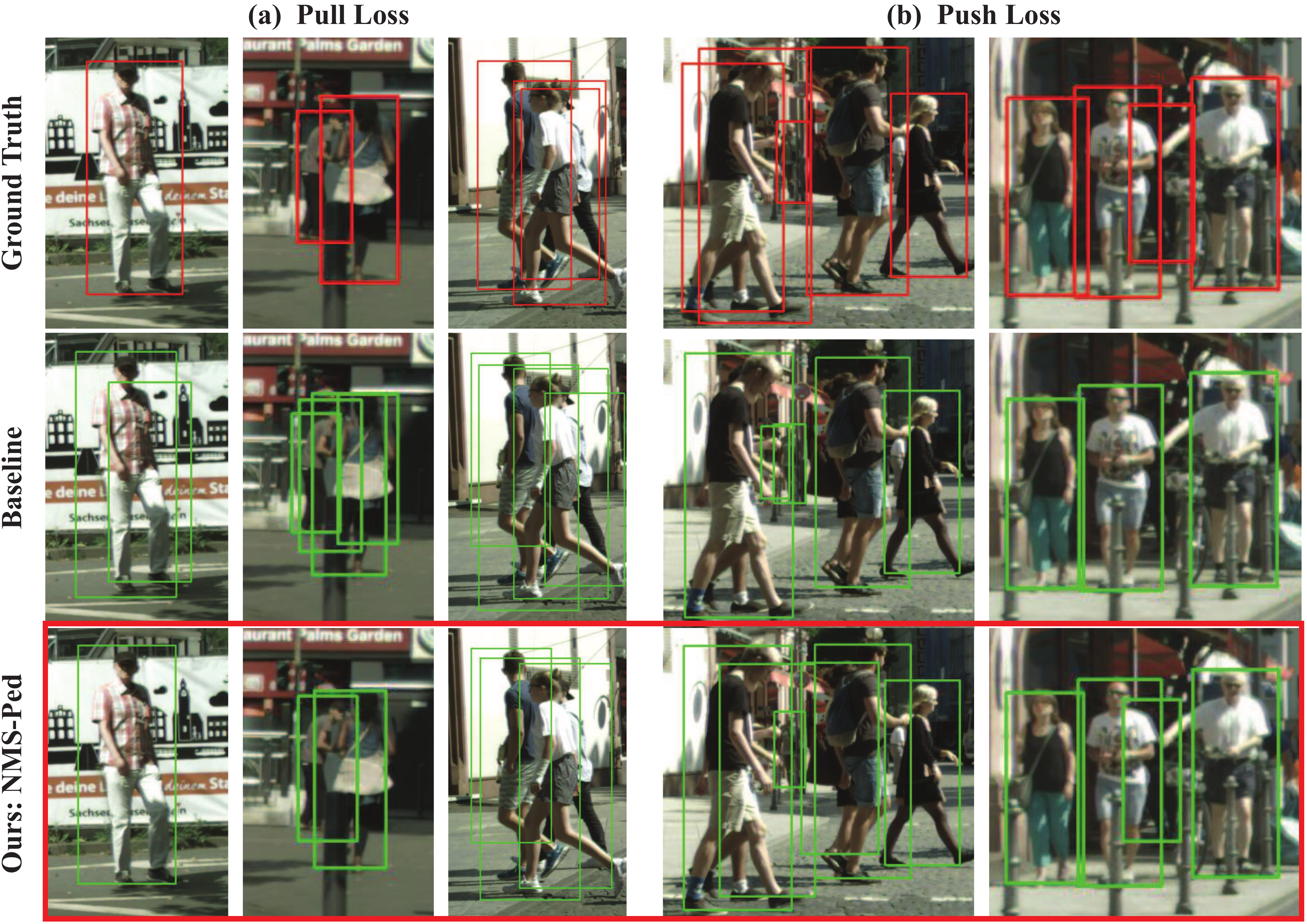}
\vspace{-2mm}
\caption{\textbf{Comparison between the cases with/without using pull/push loss}. Green bounding boxes are predicted pedestrians whose score is greater than $0.8$ and red bounding boxes are ground truth. Our pull loss effectively suppresses FP in both sparse scenes and crowded scenes (left three columns), yielding higher precision. Our push loss  robustly handles occlusions (right two columns), yielding higher recall.}
\label{visible_city_baseline_compare}
\vspace{-2mm}
\end{figure}

\begin{table}[t]
\setlength{\abovecaptionskip}{2mm}
\setlength{\belowcaptionskip}{2mm}
\caption{\textbf{Comparison of our NMS-Ped with the baseline on CityPersons}.}
\vspace{-1mm}
\begin{center}
\setlength{\tabcolsep}{10mm}{
\begin{tabular}{lccc}
\hline
Method&MR\\
\hline
\hline
baseline&$11.20\%$\\
baseline + pull loss&$10.58\%$\\
baseline + push loss&$10.61\%$\\
\textbf{NMS-Ped}&\textbf{10.08\%}\\
\hline
\end{tabular}}
\end{center}
\vspace{-2mm}
\label{table_ablation_baseline_compare}
\end{table}

\begin{table}[t]
\setlength{\abovecaptionskip}{2mm}
\setlength{\belowcaptionskip}{2mm}
\caption{\textbf{Comparison on different thresholds $N_{t}$ of NMS-Loss on CityPersons}.}
\vspace{-1mm}
\begin{center}
\setlength{\tabcolsep}{3mm}{
\begin{tabular}{l|c|c|c|c}
\hline
$N_{t}$&$0.4$&$0.45$&$\textbf{0.5}$&$0.55$ \\
\hline
MR&$10.76\%$&$10.66\%$&$\textbf{10.08\%}$&$10.67\%$ \\
\hline
\end{tabular}}
\end{center}
\vspace{-2mm}
\label{table_threshold_compared}
\end{table}

\subsection{Ablation Studies}
We conduct experiments on CityPersons to evaluate our NMS-Loss for the reason that pedestrian in CityPersons is more crowded and challenging. There are enough complicated scenes to review effectiveness of our approach.

\noindent
\textbf{Baseline comparison.}
Tab.~\ref{table_ablation_baseline_compare} shows the performance of our baseline with separate components. When only the pull loss is used, MR can be reduced from $11.20\%$ to $10.58\%$. Fig.~\ref{visible_city_baseline_compare} shows some results corrected for using pull loss. In both sparse scenes (first column) and crowded scenes (second and third columns), our pull loss will effectively pull predictions targeting on the same ground truth close to each other. The same experiments are conducted on push loss. With the help of push loss, the MR can be reduced from $11.20\%$ to $10.61\%$. Some visible results are present in Fig.~\ref{visible_city_baseline_compare} showing the corrected predictions for using push loss. In the occlusion scenes (right two columns), push loss trained model performs more robust, even detected the unlabeled instance (fourth column). When we use the complete NMS-Loss, our NMS-Ped can be boosted from both pull loss and push loss, getting an amazing $10.08\%$ MR.

\noindent
\textbf{Experiments on hyperparameters.}
Tab.~\ref{table_threshold_compared} shows our results with different thresholds $N_{t}$ on NMS-Loss.
When $N_{t}$ is lower than evaluation metric threshold $0.5$, push loss will be activated more frequently and pull loss will not be activated  making the network produce more FPs that harms precision.
In contrast, when $N_{t}$ is higher than $0.5$, more FNs will be produced and lower recall. Our NMS-Loss performs robust with various NMS thresholds, gaining stable improvement.
When we use $N_{t}$ equivalent to the threshold $0.5$, our NMS-Loss yields the best performance.

\begin{table}[t]
\setlength{\abovecaptionskip}{2mm}
\setlength{\belowcaptionskip}{2mm}
\caption{\textbf{Comparison on CityPersons dataset}.}
\vspace{-1mm}
\begin{center}
\setlength{\tabcolsep}{5mm}{
\begin{tabular}{l|c|c}
\hline
Method&Backbone&MR\\
\hline
\hline
RepLoss~\cite{wang2018repulsion} &ResNet-50&$13.20\%$\\
OR-CNN~\cite{zhang2018occlusion} &ResNet-50&$12.80\%$\\
Adaptive-NMS~\cite{liu2019adaptive} &VGG-16&$11.90\%$\\
CSP~\cite{liu2019high} &ResNet-50&$11.00\%$\\
MGAN~\cite{pang2019mask} &VGG-16&$11.50\%$\\
R$^2$NMS~\cite{huang2020nms} &VGG-16&$11.10\%$\\
EMD-RCNN~\cite{chu2020detection} &ResNet-50&$10.70\%$\\
\hline
Our baseline&ResNet-50&$11.20\%$\\
\textbf{NMS-Ped}&\textbf{ResNet-50}&\textbf{10.08\%}\\
\hline
\end{tabular}}
\end{center}
\vspace{-2mm}
\label{table_sota_compare_citypersons}
\end{table}

\begin{table}[t]
\setlength{\abovecaptionskip}{2mm}
\setlength{\belowcaptionskip}{2mm}
\caption{\textbf{Comparisons on Caltech dataset}.}
\vspace{-1mm}
\begin{center}
\setlength{\tabcolsep}{5mm}{
\begin{tabular}{l|c|c}
\hline
Method&Backbone&MR\\
\hline
\hline
RPN+BF~\cite{zhang2016faster} &VGG-16&$9.58\%$\\
F-DNN~\cite{du2017fused} &ResNet-50&$8.65\%$\\
SDS-RCNN~\cite{brazil2017illuminating} &VGG-16&$7.36\%$\\
MGAN~\cite{pang2019mask} & VGG-16 &$6.83\%$\\
AR-Ped~\cite{brazil2019pedestrian} &VGG-16&$6.45\%$\\
SSA-CNN~\cite{zhou2019ssa} &VGG-16&6$.27\%$\\
TFAN+TDEM+PRM~\cite{wu2020temporal} & ResNet-101 & 6$.50\%$\\
W$^2$Net~\cite{luo2020whether} & ResNet-50 &6$.37\%$\\
\hline
Our baseline&ResNet-50&$6.61$\%\\
\textbf{NMS-Ped}&\textbf{ResNet-50}&\textbf{5.92\%}\\
\hline
\end{tabular}}
\end{center}
\label{table_sota_compare_caltech}
\vspace{-2mm}
\end{table}

\subsection{Comparisons with SOTA methods}
To demonstrate the effectiveness of our NMS-Loss, we compare NMS-Ped with the SOTA methods on CityPersons and Caltech.
Tab.~\ref{table_sota_compare_citypersons} presents the performance of NMS-Ped and SOTA methods on the CityPersons dataset. With the help of NMS-Loss, our method improve the MR of baseline from $11.20\%$ to $10.08\%$, better than the SOTA method EMD-RCNN~\cite{chu2020detection} (MR of $10.70\%$). Tab.~\ref{table_sota_compare_caltech} presents the performance on Caltech, the MR of NMS-Ped is $5.92\%$, better than SOTA method W$^2$Net~\cite{luo2020whether} (MR of $6.37\%$). With the help of NMS-Loss, we can obtain more than $10\%$ improvement in NMS-Ped compared with baseline. This demonstrates the effectiveness of our NMS-Loss.

\subsection{Difference to RepLoss}
We make a detailed comparison between our NMS-Loss and the RepLoss~\cite{wang2018repulsion} for the reason that both methods pull and push predictions based on their targets. There are three main differences:
(1) RepLoss is performed on all instances, while NMS-Loss is only performed on instances wrongly processed by NMS, which enables end-to-end training.
(2) RepLoss only considers regression, while the score is also used in NMS-Loss to reweight instances.
(3) In dense crowd scenarios, RepLoss pushes instances away even if their targets are originally close to each other, making the repulsion loss contradicts with the regression loss.
Instead, NMS-Loss pushes instances whose IoU with others is higher than the IoU of its corresponding ground truth boxes, which can eliminates the contradiction of RepLoss. As shown in Tab.~\ref{table_impro_compare}, our NMS-Loss not only performs better than RepLoss, but also gains higher relative improvement on CityPersons. This demonstrates that our NMS-Loss can achieve stable relative improvement (higher than $10\%$) on the widely used datasets.


\begin{table}[t]
\small
\caption{\textbf{Comparison between RepLoss and NMS-Loss on the CityPersons}. We use $MR_b$, $MR$, $MR_i$, $MR_r$ to represent the $MR$ of baseline model, $MR$ of complete model, $MR$ of the improvement and relative improvement based on the baseline, respectively.}
\vspace{-1mm}
\begin{center}
\setlength{\tabcolsep}{2mm}{
\begin{tabular}{l|ccccc}
\hline
Method& Backbone & $MR_b\downarrow$ & $MR\downarrow$ &$MR_i\uparrow$&$MR_r\uparrow$\\
\hline
\hline
RepLoss & ResNet-50 & $14.6\%$  & $13.2\%$  & $1.4\%$  & $9.59\%$\\
NMS-Ped & ResNet-50 & $11.2\%$ & $10.08\%$ & $1.12\%$ & $\textbf{10.00\%}$\\
\hline
\end{tabular}}
\end{center}
\vspace{-3mm}
\label{table_impro_compare}
\end{table}



\section{Conclusion and Future Work}
In this work, we raise the problem of weak connection between training targets and evaluation metrics in the object detection. To address this, we propose the NMS-Loss which contains two components called pull loss and push loss, making the false predictions can be directly reflected on loss functions. With the help of NMS-Loss, the model can be trained with NMS end-to-end and pay more attention to the false predictions caused by NMS. Our NMS-Loss can be easily incorporated into network, which does not introduce any parameters nor runtime cost. NMS-Loss is only suitable for single class object detection, in the future, we will extend our NMS-Loss to other tasks by further considering object classes in generic detections.

\balance

\bibliographystyle{ACM-Reference-Format}

\begin{thebibliography}{39}


\ifx \showCODEN    \undefined \def \showCODEN     #1{\unskip}     \fi
\ifx \showDOI      \undefined \def \showDOI       #1{#1}\fi
\ifx \showISBNx    \undefined \def \showISBNx     #1{\unskip}     \fi
\ifx \showISBNxiii \undefined \def \showISBNxiii  #1{\unskip}     \fi
\ifx \showISSN     \undefined \def \showISSN      #1{\unskip}     \fi
\ifx \showLCCN     \undefined \def \showLCCN      #1{\unskip}     \fi
\ifx \shownote     \undefined \def \shownote      #1{#1}          \fi
\ifx \showarticletitle \undefined \def \showarticletitle #1{#1}   \fi
\ifx \showURL      \undefined \def \showURL       {\relax}        \fi
\providecommand\bibfield[2]{#2}
\providecommand\bibinfo[2]{#2}
\providecommand\natexlab[1]{#1}
\providecommand\showeprint[2][]{arXiv:#2}

\bibitem[\protect\citeauthoryear{Bodla, Singh, Chellappa, and Davis}{Bodla
  et~al\mbox{.}}{2017}]%
        {bodla2017soft}
\bibfield{author}{\bibinfo{person}{Navaneeth Bodla}, \bibinfo{person}{Bharat
  Singh}, \bibinfo{person}{Rama Chellappa}, {and} \bibinfo{person}{Larry~S
  Davis}.} \bibinfo{year}{2017}\natexlab{}.
\newblock \showarticletitle{{Soft-NMS--I}mproving Object Detection With One
  Line of Code}. In \bibinfo{booktitle}{\emph{ICCV}}.
  \bibinfo{pages}{5561--5569}.
\newblock


\bibitem[\protect\citeauthoryear{Brazil and Liu}{Brazil and Liu}{2019}]%
        {brazil2019pedestrian}
\bibfield{author}{\bibinfo{person}{Garrick Brazil} {and}
  \bibinfo{person}{Xiaoming Liu}.} \bibinfo{year}{2019}\natexlab{}.
\newblock \showarticletitle{Pedestrian Detection with Autoregressive Network
  Phases}. In \bibinfo{booktitle}{\emph{CVPR}}. \bibinfo{pages}{7231--7240}.
\newblock


\bibitem[\protect\citeauthoryear{Brazil, Yin, and Liu}{Brazil
  et~al\mbox{.}}{2017}]%
        {brazil2017illuminating}
\bibfield{author}{\bibinfo{person}{Garrick Brazil}, \bibinfo{person}{Xi Yin},
  {and} \bibinfo{person}{Xiaoming Liu}.} \bibinfo{year}{2017}\natexlab{}.
\newblock \showarticletitle{Illuminating pedestrians via simultaneous detection
  \& segmentation}. In \bibinfo{booktitle}{\emph{ICCV}}.
  \bibinfo{pages}{4950--4959}.
\newblock


\bibitem[\protect\citeauthoryear{Cai, Fan, Feris, and Vasconcelos}{Cai
  et~al\mbox{.}}{2016}]%
        {cai2016unified}
\bibfield{author}{\bibinfo{person}{Zhaowei Cai}, \bibinfo{person}{Quanfu Fan},
  \bibinfo{person}{Rogerio~S Feris}, {and} \bibinfo{person}{Nuno Vasconcelos}.}
  \bibinfo{year}{2016}\natexlab{}.
\newblock \showarticletitle{A unified multi-scale deep convolutional neural
  network for fast object detection}. In \bibinfo{booktitle}{\emph{ECCV}}.
  Springer, \bibinfo{pages}{354--370}.
\newblock


\bibitem[\protect\citeauthoryear{Chi, Zhang, Xing, Lei, Li, and Zou}{Chi
  et~al\mbox{.}}{2020a}]%
        {chi2019relational}
\bibfield{author}{\bibinfo{person}{Cheng Chi}, \bibinfo{person}{Shifeng Zhang},
  \bibinfo{person}{Junliang Xing}, \bibinfo{person}{Zhen Lei},
  \bibinfo{person}{Stan~Z Li}, {and} \bibinfo{person}{Xudong Zou}.}
  \bibinfo{year}{2020}\natexlab{a}.
\newblock \showarticletitle{Relational learning for joint head and human
  detection}. In \bibinfo{booktitle}{\emph{AAAI}}.
  \bibinfo{pages}{10647--10654}.
\newblock


\bibitem[\protect\citeauthoryear{Chi, Zhang, Xing, Lei, Li, Zou,
  et~al\mbox{.}}{Chi et~al\mbox{.}}{2020b}]%
        {chi2020pedhunter}
\bibfield{author}{\bibinfo{person}{Cheng Chi}, \bibinfo{person}{Shifeng Zhang},
  \bibinfo{person}{Junliang Xing}, \bibinfo{person}{Zhen Lei},
  \bibinfo{person}{Stan~Z Li}, \bibinfo{person}{Xudong Zou}, {et~al\mbox{.}}}
  \bibinfo{year}{2020}\natexlab{b}.
\newblock \showarticletitle{PedHunter: Occlusion Robust Pedestrian Detector in
  Crowded Scenes.}. In \bibinfo{booktitle}{\emph{AAAI}}.
  \bibinfo{pages}{10639--10646}.
\newblock


\bibitem[\protect\citeauthoryear{Chu, Zheng, Zhang, and Sun}{Chu
  et~al\mbox{.}}{2020}]%
        {chu2020detection}
\bibfield{author}{\bibinfo{person}{Xuangeng Chu}, \bibinfo{person}{Anlin
  Zheng}, \bibinfo{person}{Xiangyu Zhang}, {and} \bibinfo{person}{Jian Sun}.}
  \bibinfo{year}{2020}\natexlab{}.
\newblock \showarticletitle{Detection in Crowded Scenes: One Proposal, Multiple
  Predictions}. In \bibinfo{booktitle}{\emph{CVPR}}.
  \bibinfo{pages}{12214--12223}.
\newblock


\bibitem[\protect\citeauthoryear{Cortes and Vapnik}{Cortes and Vapnik}{1995}]%
        {cortes1995support}
\bibfield{author}{\bibinfo{person}{Corinna Cortes} {and}
  \bibinfo{person}{Vladimir Vapnik}.} \bibinfo{year}{1995}\natexlab{}.
\newblock \showarticletitle{Support-vector networks}.
\newblock \bibinfo{journal}{\emph{Machine learning}} \bibinfo{volume}{20},
  \bibinfo{number}{3} (\bibinfo{year}{1995}), \bibinfo{pages}{273--297}.
\newblock


\bibitem[\protect\citeauthoryear{Dalal and Triggs}{Dalal and Triggs}{2005}]%
        {dalal2005histograms}
\bibfield{author}{\bibinfo{person}{Navneet Dalal} {and} \bibinfo{person}{Bill
  Triggs}.} \bibinfo{year}{2005}\natexlab{}.
\newblock \showarticletitle{Histograms of oriented gradients for human
  detection}. In \bibinfo{booktitle}{\emph{CVPR}}. \bibinfo{pages}{886--893}.
\newblock


\bibitem[\protect\citeauthoryear{Doll{\'a}r, Appel, Belongie, and
  Perona}{Doll{\'a}r et~al\mbox{.}}{2014}]%
        {dollar2014fast}
\bibfield{author}{\bibinfo{person}{Piotr Doll{\'a}r}, \bibinfo{person}{Ron
  Appel}, \bibinfo{person}{Serge Belongie}, {and} \bibinfo{person}{Pietro
  Perona}.} \bibinfo{year}{2014}\natexlab{}.
\newblock \showarticletitle{Fast feature pyramids for object detection}.
\newblock \bibinfo{journal}{\emph{PAMI}} \bibinfo{volume}{36},
  \bibinfo{number}{8} (\bibinfo{year}{2014}), \bibinfo{pages}{1532--1545}.
\newblock


\bibitem[\protect\citeauthoryear{Doll{\'a}r, Tu, Perona, and
  Belongie}{Doll{\'a}r et~al\mbox{.}}{2009}]%
        {dollar2009integral}
\bibfield{author}{\bibinfo{person}{Piotr Doll{\'a}r}, \bibinfo{person}{Zhuowen
  Tu}, \bibinfo{person}{Pietro Perona}, {and} \bibinfo{person}{Serge
  Belongie}.} \bibinfo{year}{2009}\natexlab{}.
\newblock \showarticletitle{Integral channel features}. In
  \bibinfo{booktitle}{\emph{BMVC}}.
\newblock


\bibitem[\protect\citeauthoryear{Dollar, Wojek, Schiele, and Perona}{Dollar
  et~al\mbox{.}}{2009}]%
        {dollar2009pedestrian}
\bibfield{author}{\bibinfo{person}{P Dollar}, \bibinfo{person}{C Wojek},
  \bibinfo{person}{B Schiele}, {and} \bibinfo{person}{P Perona}.}
  \bibinfo{year}{2009}\natexlab{}.
\newblock \showarticletitle{Pedestrian detection: A benchmark}. In
  \bibinfo{booktitle}{\emph{CVPR}}. \bibinfo{pages}{304--311}.
\newblock


\bibitem[\protect\citeauthoryear{Dollar, Wojek, Schiele, and Perona}{Dollar
  et~al\mbox{.}}{2011}]%
        {dollar2011pedestrian}
\bibfield{author}{\bibinfo{person}{Piotr Dollar}, \bibinfo{person}{Christian
  Wojek}, \bibinfo{person}{Bernt Schiele}, {and} \bibinfo{person}{Pietro
  Perona}.} \bibinfo{year}{2011}\natexlab{}.
\newblock \showarticletitle{Pedestrian detection: An evaluation of the state of
  the art}.
\newblock \bibinfo{journal}{\emph{PAMI}} \bibinfo{volume}{34},
  \bibinfo{number}{4} (\bibinfo{year}{2011}), \bibinfo{pages}{743--761}.
\newblock


\bibitem[\protect\citeauthoryear{Du, El-Khamy, Lee, and Davis}{Du
  et~al\mbox{.}}{2017}]%
        {du2017fused}
\bibfield{author}{\bibinfo{person}{Xianzhi Du}, \bibinfo{person}{Mostafa
  El-Khamy}, \bibinfo{person}{Jungwon Lee}, {and} \bibinfo{person}{Larry
  Davis}.} \bibinfo{year}{2017}\natexlab{}.
\newblock \showarticletitle{{Fused DNN}: A deep neural network fusion approach
  to fast and robust pedestrian detection}. In
  \bibinfo{booktitle}{\emph{WACV}}. IEEE, \bibinfo{pages}{953--961}.
\newblock


\bibitem[\protect\citeauthoryear{Felzenszwalb, Girshick, and
  McAllester}{Felzenszwalb et~al\mbox{.}}{2010}]%
        {felzenszwalb2010cascade}
\bibfield{author}{\bibinfo{person}{Pedro~F Felzenszwalb},
  \bibinfo{person}{Ross~B Girshick}, {and} \bibinfo{person}{David McAllester}.}
  \bibinfo{year}{2010}\natexlab{}.
\newblock \showarticletitle{Cascade object detection with deformable part
  models}. In \bibinfo{booktitle}{\emph{CVPR}}. IEEE,
  \bibinfo{pages}{2241--2248}.
\newblock


\bibitem[\protect\citeauthoryear{Felzenszwalb, Girshick, McAllester, and
  Ramanan}{Felzenszwalb et~al\mbox{.}}{2009}]%
        {felzenszwalb2009object}
\bibfield{author}{\bibinfo{person}{Pedro~F Felzenszwalb},
  \bibinfo{person}{Ross~B Girshick}, \bibinfo{person}{David McAllester}, {and}
  \bibinfo{person}{Deva Ramanan}.} \bibinfo{year}{2009}\natexlab{}.
\newblock \showarticletitle{Object detection with discriminatively trained
  part-based models}.
\newblock \bibinfo{journal}{\emph{PAMI}} \bibinfo{volume}{32},
  \bibinfo{number}{9} (\bibinfo{year}{2009}), \bibinfo{pages}{1627--1645}.
\newblock


\bibitem[\protect\citeauthoryear{He, Zhang, Ren, and Sun}{He
  et~al\mbox{.}}{2016}]%
        {he2016deep}
\bibfield{author}{\bibinfo{person}{Kaiming He}, \bibinfo{person}{Xiangyu
  Zhang}, \bibinfo{person}{Shaoqing Ren}, {and} \bibinfo{person}{Jian Sun}.}
  \bibinfo{year}{2016}\natexlab{}.
\newblock \showarticletitle{Deep residual learning for image recognition}. In
  \bibinfo{booktitle}{\emph{CVPR}}. \bibinfo{pages}{770--778}.
\newblock


\bibitem[\protect\citeauthoryear{Huang, Ge, Jie, and Yoshie}{Huang
  et~al\mbox{.}}{2020}]%
        {huang2020nms}
\bibfield{author}{\bibinfo{person}{Xin Huang}, \bibinfo{person}{Zheng Ge},
  \bibinfo{person}{Zequn Jie}, {and} \bibinfo{person}{Osamu Yoshie}.}
  \bibinfo{year}{2020}\natexlab{}.
\newblock \showarticletitle{NMS by Representative Region: Towards Crowded
  Pedestrian Detection by Proposal Pairing}. In
  \bibinfo{booktitle}{\emph{CVPR}}. \bibinfo{pages}{10750--10759}.
\newblock


\bibitem[\protect\citeauthoryear{Liu, Huang, and Wang}{Liu
  et~al\mbox{.}}{2019a}]%
        {liu2019adaptive}
\bibfield{author}{\bibinfo{person}{Songtao Liu}, \bibinfo{person}{Di Huang},
  {and} \bibinfo{person}{Yunhong Wang}.} \bibinfo{year}{2019}\natexlab{a}.
\newblock \showarticletitle{{Adaptive NMS}: Refining Pedestrian Detection in a
  Crowd}. In \bibinfo{booktitle}{\emph{CVPR}}. \bibinfo{pages}{6459--6468}.
\newblock


\bibitem[\protect\citeauthoryear{Liu, Liao, Hu, Liang, and Chen}{Liu
  et~al\mbox{.}}{2018}]%
        {liu2018learning}
\bibfield{author}{\bibinfo{person}{Wei Liu}, \bibinfo{person}{Shengcai Liao},
  \bibinfo{person}{Weidong Hu}, \bibinfo{person}{Xuezhi Liang}, {and}
  \bibinfo{person}{Xiao Chen}.} \bibinfo{year}{2018}\natexlab{}.
\newblock \showarticletitle{Learning efficient single-stage pedestrian
  detectors by asymptotic localization fitting}. In
  \bibinfo{booktitle}{\emph{ECCV}}. \bibinfo{pages}{618--634}.
\newblock


\bibitem[\protect\citeauthoryear{Liu, Liao, Ren, Hu, and Yu}{Liu
  et~al\mbox{.}}{2019b}]%
        {liu2019high}
\bibfield{author}{\bibinfo{person}{Wei Liu}, \bibinfo{person}{Shengcai Liao},
  \bibinfo{person}{Weiqiang Ren}, \bibinfo{person}{Weidong Hu}, {and}
  \bibinfo{person}{Yinan Yu}.} \bibinfo{year}{2019}\natexlab{b}.
\newblock \showarticletitle{High-level Semantic Feature Detection: A New
  Perspective for Pedestrian Detection}. In \bibinfo{booktitle}{\emph{CVPR}}.
  \bibinfo{pages}{5187--5196}.
\newblock


\bibitem[\protect\citeauthoryear{Luo, Zhang, Zhao, Zhou, and Sun}{Luo
  et~al\mbox{.}}{2020}]%
        {luo2020whether}
\bibfield{author}{\bibinfo{person}{Yan Luo}, \bibinfo{person}{Chongyang Zhang},
  \bibinfo{person}{Muming Zhao}, \bibinfo{person}{Hao Zhou}, {and}
  \bibinfo{person}{Jun Sun}.} \bibinfo{year}{2020}\natexlab{}.
\newblock \showarticletitle{Where, What, Whether: Multi-Modal Learning Meets
  Pedestrian Detection}. In \bibinfo{booktitle}{\emph{{CVPR}}}.
  \bibinfo{pages}{14065--14073}.
\newblock


\bibitem[\protect\citeauthoryear{Mao, Xiao, Jiang, and Cao}{Mao
  et~al\mbox{.}}{2017}]%
        {mao2017can}
\bibfield{author}{\bibinfo{person}{Jiayuan Mao}, \bibinfo{person}{Tete Xiao},
  \bibinfo{person}{Yuning Jiang}, {and} \bibinfo{person}{Zhimin Cao}.}
  \bibinfo{year}{2017}\natexlab{}.
\newblock \showarticletitle{What can help pedestrian detection?}. In
  \bibinfo{booktitle}{\emph{CVPR}}. \bibinfo{pages}{3127--3136}.
\newblock


\bibitem[\protect\citeauthoryear{Nam, Doll{\'a}r, and Han}{Nam
  et~al\mbox{.}}{2014}]%
        {nam2014local}
\bibfield{author}{\bibinfo{person}{Woonhyun Nam}, \bibinfo{person}{Piotr
  Doll{\'a}r}, {and} \bibinfo{person}{Joon~Hee Han}.}
  \bibinfo{year}{2014}\natexlab{}.
\newblock \showarticletitle{Local decorrelation for improved pedestrian
  detection}. In \bibinfo{booktitle}{\emph{NIPS}}. \bibinfo{pages}{424--432}.
\newblock


\bibitem[\protect\citeauthoryear{Pang, Xie, Khan, Anwer, Khan, and Shao}{Pang
  et~al\mbox{.}}{2019}]%
        {pang2019mask}
\bibfield{author}{\bibinfo{person}{Yanwei Pang}, \bibinfo{person}{Jin Xie},
  \bibinfo{person}{Muhammad~Haris Khan}, \bibinfo{person}{Rao~Muhammad Anwer},
  \bibinfo{person}{Fahad~Shahbaz Khan}, {and} \bibinfo{person}{Ling Shao}.}
  \bibinfo{year}{2019}\natexlab{}.
\newblock \showarticletitle{Mask-Guided Attention Network for Occluded
  Pedestrian Detection}. In \bibinfo{booktitle}{\emph{ICCV}}.
  \bibinfo{pages}{4967--4975}.
\newblock


\bibitem[\protect\citeauthoryear{Paszke, Gross, Chintala, Chanan, Yang, DeVito,
  Lin, Desmaison, Antiga, and Lerer}{Paszke et~al\mbox{.}}{2017}]%
        {paszke2017automatic}
\bibfield{author}{\bibinfo{person}{Adam Paszke}, \bibinfo{person}{Sam Gross},
  \bibinfo{person}{Soumith Chintala}, \bibinfo{person}{Gregory Chanan},
  \bibinfo{person}{Edward Yang}, \bibinfo{person}{Zachary DeVito},
  \bibinfo{person}{Zeming Lin}, \bibinfo{person}{Alban Desmaison},
  \bibinfo{person}{Luca Antiga}, {and} \bibinfo{person}{Adam Lerer}.}
  \bibinfo{year}{2017}\natexlab{}.
\newblock \showarticletitle{Automatic differentiation in PyTorch}.
\newblock  (\bibinfo{year}{2017}).
\newblock


\bibitem[\protect\citeauthoryear{Ren, Chen, Liu, Sun, Pang, Yan, Tai, and
  Xu}{Ren et~al\mbox{.}}{2017}]%
        {ren2017accurate}
\bibfield{author}{\bibinfo{person}{Jimmy Ren}, \bibinfo{person}{Xiaohao Chen},
  \bibinfo{person}{Jianbo Liu}, \bibinfo{person}{Wenxiu Sun},
  \bibinfo{person}{Jiahao Pang}, \bibinfo{person}{Qiong Yan},
  \bibinfo{person}{Yu-Wing Tai}, {and} \bibinfo{person}{Li Xu}.}
  \bibinfo{year}{2017}\natexlab{}.
\newblock \showarticletitle{Accurate single stage detector using recurrent
  rolling convolution}. In \bibinfo{booktitle}{\emph{CVPR}}.
  \bibinfo{pages}{5420--5428}.
\newblock


\bibitem[\protect\citeauthoryear{Ren, He, Girshick, and Sun}{Ren
  et~al\mbox{.}}{2015}]%
        {ren2015faster}
\bibfield{author}{\bibinfo{person}{Shaoqing Ren}, \bibinfo{person}{Kaiming He},
  \bibinfo{person}{Ross Girshick}, {and} \bibinfo{person}{Jian Sun}.}
  \bibinfo{year}{2015}\natexlab{}.
\newblock \showarticletitle{{Faster R-CNN}: Towards real-time object detection
  with region proposal networks}. In \bibinfo{booktitle}{\emph{NIPS}}.
  \bibinfo{pages}{91--99}.
\newblock


\bibitem[\protect\citeauthoryear{Simonyan and Zisserman}{Simonyan and
  Zisserman}{2014}]%
        {simonyan2014very}
\bibfield{author}{\bibinfo{person}{Karen Simonyan} {and}
  \bibinfo{person}{Andrew Zisserman}.} \bibinfo{year}{2014}\natexlab{}.
\newblock \showarticletitle{Very deep convolutional networks for large-scale
  image recognition}.
\newblock \bibinfo{journal}{\emph{arXiv preprint arXiv:1409.1556}}
  (\bibinfo{year}{2014}).
\newblock


\bibitem[\protect\citeauthoryear{Viola and Jones}{Viola and Jones}{2001}]%
        {viola2001rapid}
\bibfield{author}{\bibinfo{person}{Paul Viola} {and} \bibinfo{person}{Michael
  Jones}.} \bibinfo{year}{2001}\natexlab{}.
\newblock \showarticletitle{Rapid object detection using a boosted cascade of
  simple features}. In \bibinfo{booktitle}{\emph{CVPR}}. \bibinfo{pages}{I--I}.
\newblock


\bibitem[\protect\citeauthoryear{Wang, Xiao, Jiang, Shao, Sun, and Shen}{Wang
  et~al\mbox{.}}{2018}]%
        {wang2018repulsion}
\bibfield{author}{\bibinfo{person}{Xinlong Wang}, \bibinfo{person}{Tete Xiao},
  \bibinfo{person}{Yuning Jiang}, \bibinfo{person}{Shuai Shao},
  \bibinfo{person}{Jian Sun}, {and} \bibinfo{person}{Chunhua Shen}.}
  \bibinfo{year}{2018}\natexlab{}.
\newblock \showarticletitle{Repulsion loss: Detecting pedestrians in a crowd}.
  In \bibinfo{booktitle}{\emph{CVPR}}. \bibinfo{pages}{7774--7783}.
\newblock


\bibitem[\protect\citeauthoryear{Wu, Zhou, Yang, Zhang, Li, and Yuan}{Wu
  et~al\mbox{.}}{2020}]%
        {wu2020temporal}
\bibfield{author}{\bibinfo{person}{Jialian Wu}, \bibinfo{person}{Chunluan
  Zhou}, \bibinfo{person}{Ming Yang}, \bibinfo{person}{Qian Zhang},
  \bibinfo{person}{Yuan Li}, {and} \bibinfo{person}{Junsong Yuan}.}
  \bibinfo{year}{2020}\natexlab{}.
\newblock \showarticletitle{Temporal-Context Enhanced Detection of Heavily
  Occluded Pedestrians}. In \bibinfo{booktitle}{\emph{CVPR}}.
  \bibinfo{pages}{13430--13439}.
\newblock


\bibitem[\protect\citeauthoryear{Xu, Ramos, V{\'a}zquez, and L{\'o}pez}{Xu
  et~al\mbox{.}}{2014}]%
        {xu2014domain}
\bibfield{author}{\bibinfo{person}{Jiaolong Xu}, \bibinfo{person}{Sebastian
  Ramos}, \bibinfo{person}{David V{\'a}zquez}, {and} \bibinfo{person}{Antonio~M
  L{\'o}pez}.} \bibinfo{year}{2014}\natexlab{}.
\newblock \showarticletitle{Domain adaptation of deformable part-based models}.
\newblock \bibinfo{journal}{\emph{PAMI}} \bibinfo{volume}{36},
  \bibinfo{number}{12} (\bibinfo{year}{2014}), \bibinfo{pages}{2367--2380}.
\newblock


\bibitem[\protect\citeauthoryear{Yan, Lei, Wen, and Li}{Yan
  et~al\mbox{.}}{2014}]%
        {yan2014fastest}
\bibfield{author}{\bibinfo{person}{Junjie Yan}, \bibinfo{person}{Zhen Lei},
  \bibinfo{person}{Longyin Wen}, {and} \bibinfo{person}{Stan~Z Li}.}
  \bibinfo{year}{2014}\natexlab{}.
\newblock \showarticletitle{The fastest deformable part model for object
  detection}. In \bibinfo{booktitle}{\emph{CVPR}}. \bibinfo{pages}{2497--2504}.
\newblock


\bibitem[\protect\citeauthoryear{Zhang, Lin, Liang, and He}{Zhang
  et~al\mbox{.}}{2016}]%
        {zhang2016faster}
\bibfield{author}{\bibinfo{person}{Liliang Zhang}, \bibinfo{person}{Liang Lin},
  \bibinfo{person}{Xiaodan Liang}, {and} \bibinfo{person}{Kaiming He}.}
  \bibinfo{year}{2016}\natexlab{}.
\newblock \showarticletitle{Is {Faster R-CNN} doing well for pedestrian
  detection?}. In \bibinfo{booktitle}{\emph{ECCV}}. Springer,
  \bibinfo{pages}{443--457}.
\newblock


\bibitem[\protect\citeauthoryear{Zhang, Benenson, and Schiele}{Zhang
  et~al\mbox{.}}{2017}]%
        {zhang2017citypersons}
\bibfield{author}{\bibinfo{person}{Shanshan Zhang}, \bibinfo{person}{Rodrigo
  Benenson}, {and} \bibinfo{person}{Bernt Schiele}.}
  \bibinfo{year}{2017}\natexlab{}.
\newblock \showarticletitle{Citypersons: A diverse dataset for pedestrian
  detection}. In \bibinfo{booktitle}{\emph{CVPR}}. \bibinfo{pages}{3213--3221}.
\newblock


\bibitem[\protect\citeauthoryear{Zhang, Benenson, Schiele, et~al\mbox{.}}{Zhang
  et~al\mbox{.}}{2015}]%
        {zhang2015filtered}
\bibfield{author}{\bibinfo{person}{Shanshan Zhang}, \bibinfo{person}{Rodrigo
  Benenson}, \bibinfo{person}{Bernt Schiele}, {et~al\mbox{.}}}
  \bibinfo{year}{2015}\natexlab{}.
\newblock \showarticletitle{Filtered channel features for pedestrian
  detection.}. In \bibinfo{booktitle}{\emph{CVPR}}, Vol.~\bibinfo{volume}{1}.
  \bibinfo{pages}{4}.
\newblock


\bibitem[\protect\citeauthoryear{Zhang, Wen, Bian, Lei, and Li}{Zhang
  et~al\mbox{.}}{2018}]%
        {zhang2018occlusion}
\bibfield{author}{\bibinfo{person}{Shifeng Zhang}, \bibinfo{person}{Longyin
  Wen}, \bibinfo{person}{Xiao Bian}, \bibinfo{person}{Zhen Lei}, {and}
  \bibinfo{person}{Stan~Z Li}.} \bibinfo{year}{2018}\natexlab{}.
\newblock \showarticletitle{Occlusion-aware {R-CNN}: detecting pedestrians in a
  crowd}. In \bibinfo{booktitle}{\emph{ECCV}}. \bibinfo{pages}{637--653}.
\newblock


\bibitem[\protect\citeauthoryear{Zhou, Wu, and Lam}{Zhou et~al\mbox{.}}{2019}]%
        {zhou2019ssa}
\bibfield{author}{\bibinfo{person}{Chengju Zhou}, \bibinfo{person}{Meiqing Wu},
  {and} \bibinfo{person}{Siew-Kei Lam}.} \bibinfo{year}{2019}\natexlab{}.
\newblock \showarticletitle{{SSA-CNN}: Semantic Self-Attention CNN for
  Pedestrian Detection}.
\newblock \bibinfo{journal}{\emph{arXiv preprint arXiv:1902.09080}}
  (\bibinfo{year}{2019}).
\newblock


\end{thebibliography}

\end{document}